\documentclass[letterpaper]{article} 
\usepackage{aaai2026}  
\usepackage{times}  
\usepackage{helvet}  
\usepackage{courier}  
\usepackage[hyphens]{url}  
\usepackage{bibentry}
\usepackage{graphicx}
\usepackage{graphicx} 
\usepackage{natbib}  
\usepackage{caption} 
\usepackage{xcolor}
\usepackage{algorithm}
\usepackage{algorithmic}
\usepackage{amsmath} 
\usepackage{amsfonts}
\usepackage{booktabs}

\usepackage{adjustbox}
\usepackage{makecell}

\usepackage{newfloat}
\usepackage{listings}
\DeclareCaptionStyle{ruled}{labelfont=normalfont,labelsep=colon,strut=off} 
\floatstyle{ruled}
\newfloat{listing}{tb}{lst}{}
\floatname{listing}{Listing}
\makeatletter
\ifdefined\copyright@text\else
  \def\copyright@text{Copyright \copyright\ 2025, All rights reserved.}
\fi
\makeatother

\begin{document}
\title{Governance-as-a-Service: A Multi-Agent Framework for AI System Compliance and Policy Enforcement}
\author{
    Suyash Gaurav\equalcontrib\textsuperscript{\rm 1},
    Jukka Heikkonen\textsuperscript{\rm 2},
    Jatin Chaudhary\textsuperscript{\#2}
}
\affiliations{
    \textsuperscript{\rm 1}Turku School of Economics, University of Turku \\
    \textsuperscript{\rm 2}Department of Digital Business \& Innovation, Tokyo International University \\
    \textsuperscript{\rm 3}Department of Computing, University of Turku \\


    $^\#$ jatin.chaudhary@utu.fi
    
%
}

\maketitle 

\begin{abstract}
As AI systems evolve into distributed, agentic ecosystems capable of autonomous task execution, asynchronous reasoning, and multi-agent coordination the absence of scalable, decoupled governance remains a structural liability. Existing oversight mechanisms are typically reactive, hardcoded, or embedded within agent architectures, rendering them brittle, non-auditable, and difficult to generalize across heterogeneous deployments.
We propose Governance-as-a-Service (GaaS): a modular, policy-driven enforcement layer that governs agent outputs at runtime without modifying internal model logic or assuming agent cooperation. GaaS operates through declarative rule sets and a Trust Factor mechanism that scores agents based on longitudinal compliance and severity-aware violation history. It supports coercive, normative, and adaptive interventions, allowing for graduated enforcement and per-agent trust modulation.
To empirically evaluate GaaS, we design three simulation regimes using open-source language models (LLaMA3, Qwen3, DeepSeek-R1) across two critical domains: content generation and financial decision-making. In the baseline, agents operate without governance; in the second, GaaS is deployed as an enforcement layer; in the third, adversarial agents are introduced to probe robustness. All agent actions are intercepted, evaluated, and logged for downstream analysis. Results indicate that GaaS consistently blocks or redirects high-risk behaviors while preserving agentic throughput. Trust scores evolve in alignment with rule compliance, demonstrating the system’s ability to isolate, penalize, and adapt to untrustworthy components within complex multi-agent systems. By treating governance as a runtime service on par with compute, storage, or memory GaaS establishes a foundation for infrastructure-level alignment in unregulated, interoperable agent ecosystems. It does not teach agents ethics; it enforces them.

\end{abstract}


\section{Introduction}
The emergence of AI agents marks a significant evolution in machine learning transforming theoretical constructs into modular, production-grade systems capable of multi-layered task execution, long-horizon planning, and hierarchical reasoning \cite{russell2015research, hughes2025ai, acharya2025agentic}. These agents now write public-facing content, execute financial transfers, and even control infrastructure-level actions with minimal human intervention. This newfound autonomy is often framed as a strength, but it also introduces a structural weakness: governance in agentic environments becomes deeply entangled with the agents themselves because of the decentralized nature of these agentic environment \cite{russell2015research}. With each additional decision delegated to an agent (who’s brain is an opaque large language model), ethical control shifts further from the user and into unregulated inference space.

Consider an enterprise-level orchestration environment composed of dozens of autonomous agents, a planner, multiple retrieval chains, a budget allocator, an investment executor, a legal compliance checker, and an external messaging agent. Many of these are built on open-source LLMs (due to practical constraints and customizations) or API-lightweight components that lack the safety architecture of commercial offerings such as GPT-4 or Claude. These models are not fine-tuned with constitutional AI, nor equipped with RLHF guardrails. Yet they interact, hand off results, and issue commands, sometimes downstream to production systems. In such settings, one hallucinated command, one ethically problematic statement, or one misaligned escalation path can violate regulatory requirements, erode user trust, or trigger reputational harm.

To address this risk, we propose Governance-as-a-Service (GaaS): a plug-and-play governance protocol that inserts a modular enforcement layer between agentic environment and the users. GaaS decouples behavioral enforcement from agent internals operating as a runtime service that governs through observable outputs, without requiring access to weights, prompts, or internal memory states. It supports multiple enforcement modes: coercive (blocking), normative (warning), and adaptive (escalating based on trust history), and delivers per-agent enforcement using domain-adaptive precision and a severity-aware penalization strategy.
GaaS is designed to be provisioned like infrastructure, as ubiquitous and decoupled as storage or compute. Its enforcement engine uses declarative rules defined in JSON to check agent actions against policies, issuing responses (e.g., allow, warn, block) while maintaining a trust signal for each agent over time. This allows system designers to not only prevent harm, but to measure alignment, diagnose risky agents, and adapt enforcement based on longitudinal behavior. Importantly, GaaS requires no agent cooperation and can govern unmodified models operating in black-box mode.

We demonstrate GaaS in two high-risk domains: (1) content generation, where unfiltered LLM outputs can express unethical or biased viewpoints, and (2) financial transaction automation, where violations of risk policies can lead to capital loss. In both settings, GaaS dynamically blocks unsafe actions, computes trust scores based on per-agent rule violations, and enables real-time observability of agent compliance. Across three open-source LLMs DeepSeek-R1, Llama-3, and Qwen-3, we show how GaaS governs heterogeneous agents without disrupting system liveness or functionality.
GaaS reframes governance not as a static alignment artifact, but as an enforceable runtime contract, a mechanism for making misbehavior non-executable \cite{bratman1987intention, gabriel2020artificial, arnold2017fat}. By operationalizing governance as a measurable, auditable, and adaptive system service, GaaS lays foundational groundwork for scalable AI oversight, especially in decentralized, open-source agentic ecosystems.

\noindent\textbf{Our key contributions are as follows:}
\begin{itemize}
    \item We propose \textbf{Governance-as-a-Service (GaaS)}, a protocol that governs agentic environment by using a modular enforcement layer between agentic environment and their users to enable real-time policy enforcement.

    \item We introduce an \textbf{extensible policy engine} that supports coercive and normative governance modes, along with adaptive escalation based on violation history.

    \item We construct a \textbf{simulation framework and evaluation pipeline} to assess agent compliance over time, log enforcement decisions, and support future benchmarking.

    \item We reframe AI governance as a \textbf{provisioned runtime service}, establishing a measurement protocol suitable for large-scale multi-agent simulations.

\end{itemize}

\vspace{1em}
\begin{figure*}[t]
  \centering
  \includegraphics[width=0.8\textwidth]{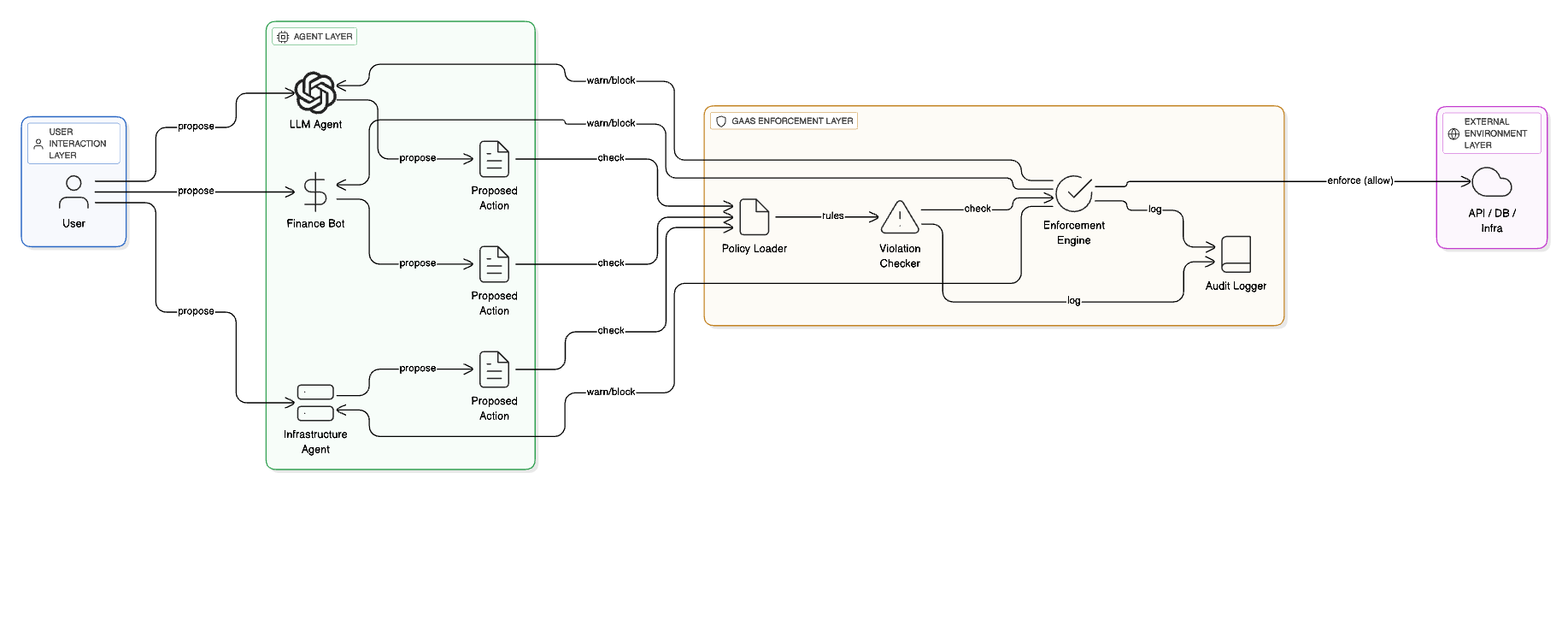}

    \caption{The GaaS architecture separates agent cognition from governance enforcement. Agents propose actions based on internal logic, which are intercepted by the GaaS layer. The enforcement engine(Trust Factor) evaluates these actions against a set of declarative policies using contextual information and pattern-matching rules. Depending on the violation outcome, the system either allows the action to proceed to the external user or returns a warning or block to the agent. All enforcement decisions are logged for traceability and adaptive escalation.}

  \label{fig:gaas-arch}
\end{figure*}

\section{Related Work}

The predominant paradigms for AI governance bifurcate into two broad trajectories: \emph{internal alignment} mechanisms that modulate agent behavior via learned preferences, and \emph{external containment} mechanisms such as filters or firewalls that constrain outputs post hoc. Both approaches are intrinsically coupled to the agent’s internal logic, limiting their generalizability and operational scalability. In contrast, our work reconceptualizes governance as a form of \textit{runtime infrastructure}, provisioned independently of agent architecture, and deployed modularly as a policy enforcement layer. This decoupling enables real-time governance that is both model-agnostic and operationally composable. Below, we situate our contribution within the broader literature.

\paragraph{Value Learning and AI Alignment:}
The AI alignment literature is classically concerned with ensuring that autonomous agents behave in ways that are consistent with human goals and ethical norms. As noted by \cite{saffarizadeh2024relationship}, alignment comprises both a technical component encoding utility functions or behavioral targets into agents—and a normative component, determining which values ought to be encoded. Techniques in this space include cooperative inverse reinforcement learning \cite{hadfield2016cooperative}, scalable preference modeling \cite{leike2018scalable}, and recursive supervision \cite{russell2015research}. While these methods have demonstrated promise, they rely on cooperative and introspectable agents, assumptions that frequently fail in open, multi-agent deployments. Our framework diverges sharply: GaaS does not seek to align internal behavior, but to constrain observable outputs through external enforcement.

\paragraph{Oversight and Corrigibility.}
Efforts in AI oversight seek to ensure post-deployment steerability via mechanisms such as corrigibility incentives, overseer modeling, and recursive auditing \cite{amodei2016concrete, christiano2020recursively}. These frameworks, however, often presume architectural entanglement—agents must be built to accept oversight. GaaS relaxes this assumption entirely: it \emph{externalizes} governance logic into a runtime policy interface that operates independently of agent cooperation, enabling enforcement even in black-box or non-corrigible agents.

\paragraph{Safety Filters and Output Firewalls.}
Lexical filters and safety firewalls constitute another class of governance tools, focused on output-level constraint via keyword matching, regular expressions, or classification-based rejection mechanisms \cite{solaiman2021process, ouyang2022training}. While effective in narrow domains, these approaches are often brittle, non-adaptive, and context-insensitive. GaaS inherits this foundational logic but generalizes it through an extensible, declarative rule engine that supports both \emph{coercive} (blocking) and \emph{normative} (warning) modes, as well as trust-based adaptive escalation grounded in historical violation patterns.

\paragraph{Policy-as-Code and Infrastructure Governance.}
Frameworks such as Open Policy Agent (OPA) \cite{opa} enable enforcement of declarative policies over access control and service boundaries. However, such systems are built for infrastructure governance—not autonomous decision-making entities. GaaS bridges this gap by preserving the declarativity of policy-as-code, while generalizing its application to domain-agnostic agent behavior and compositional multi-agent ecosystems.

\noindent In summary, GaaS does not aim to \emph{predict}, \emph{align}, or \emph{sandbox} agent behavior. Instead, it introduces a distinct paradigm: governing agentic systems through runtime modular enforcement that is transparent, interpretable, and decoupled from internal cognition. This reorientation enables scalable governance across untrusted agents and dynamic operational contexts.

\section{Methodology}

This study evaluates the Governance-as-a-Service (GaaS) protocol, a system designed to enforce compliance and manage trust in multi-agent environments. The evaluation is conducted through two distinct simulations: financial trading and collaborative AI writing. Both simulations are implemented in Python, leveraging large language models (LLMs) such as DeepSeek-R1, Llama3, and Qwen3, accessed via the Ollama client to ensure consistent and reproducible results. The GaaS framework comprises three main components: agents responsible for executing domain-specific tasks (e.g., generating trades or writing content), a governance layer that enforces compliance through a configurable rule engine, and a dynamic trust mechanism that evaluates agent reliability based on their adherence to rules over time. These components work together to simulate real-world scenarios where autonomous agents must operate within defined boundaries, providing insights into the effectiveness of GaaS in maintaining system integrity.

\subsection{System Architecture}

Governance-as-a-Service (GaaS) is designed as a modular runtime governance layer that interposes itself between autonomous agents and the external environments they seek to act upon. Its core purpose is to enforce behavioral policies in real time without requiring model retraining, internal instrumentation, or architectural cooperation from the agents it governs. This separation of control from cognition allows GaaS to operate transparently across heterogeneous, black-box agent stacks. As illustrated in Figure~\ref{fig:gaas-arch}, the system consists of three modular components: the Agent System, the GaaS Enforcement Layer, and the External Environment.

\subsubsection{Agentic System:} This module includes any autonomous agent whether LLM-based or rule-based that initiates actions. Agents are treated as untrusted by default and may differ widely in capability, purpose, and backend architecture. GaaS assumes no visibility into the model weights or training history of these agents, reflecting a realistic operational deployment scenario where agents may be sourced from open repositories or interact via lightweight APIs.

\subsubsection{GaaS Enforcement Layer:} This is the operational heart of the system. It begins with a declarative policy repository, which contains human-authored enforcement rules defined in JSON format. Rules are parsed, validated, and activated at runtime through the Policy Loader, which ensures syntactic and semantic consistency. These policies specify not only what to enforce (e.g., hate speech, hallucinated facts, financial overreach) but how to enforce it each rule is annotated with a governance mode: either coercive (block) or normative (warn).

Incoming actions from agents are intercepted and passed to the Violation Checker, which applies structured pattern-matching logic to detect rule breaches. The checker evaluates each action using contextual metadata such as agent identity, action type, accessed resources, and historical violation records. Unlike probabilistic classifiers or LLM judges, this system is deterministic, explainable, and interpretable, avoiding the opacity of neural filters.

Detected violations are sent to the Enforcement Engine, which uses the agent’s current and prior behavior to compute a Trust Factor. This trust signal is updated continuously using a severity-aware penalization strategy, allowing the system to escalate enforcement for repeat offenders. Trust scores guide decisions to allow, warn, block, or escalate enforcement. In addition to direct action, the engine also supports adaptive governance by adjusting thresholds based on violation patterns over time.

Each decision is recorded in a time-stamped audit trail that includes the triggering rule ID, agent identifier, enforcement action, and context. For critical violations, GaaS can trigger external alerts or invoke system-level callbacks, enabling integration with broader compliance or monitoring frameworks.

\subsubsection{External or User Environment:} This module represents any downstream system the agent is trying to act upon databases, web APIs, content platforms, UI interfaces, or real-world infrastructure. Importantly, only actions explicitly cleared by the Enforcement Layer are permitted to reach this environment. This structure treats governance not as a reactive afterthought but as a provisioned gatekeeper, making misbehavior computationally non-executable.

In summary, GaaS transforms governance into a first-class runtime service. Its architecture provides modular enforcement, domain-adaptive precision, and plug-and-play deployment without interfering with agent internals. By treating compliance as a programmable API contract, GaaS enables scalable, explainable, and consistent oversight across increasingly complex, decentralized agentic ecosystems.

\subsection{Governance Modes and Enforcement Logic}

The Governance-as-a-Service (GaaS) framework supports three primary governance modes: coercive, normative, and adaptive. Each mode defines a distinct enforcement philosophy and is implemented via a shared enforcement pipeline. We formalize the behavior of GaaS through an enforcement function and define how rule types, agent histories, and the trust factor influence decision outcomes.

\subsection{Trust Factor Computation}

A central component of the GaaS enforcement mechanism is the \emph{trust factor}, denoted by \( TF_a \), which serves as a dynamic, longitudinal signal representing an agent’s compliance profile. It is a scalar measure that modulates enforcement decisions by incorporating both the frequency and severity of rule violations. Formally, \( TF_a \) is computed after each agent action in accordance with Algorithm~\ref{alg:trust_factor}, and reflects a principled trade-off between behavioral transparency and policy enforcement.

Each action be it a content generation event or a financial transaction is intercepted and evaluated for potential violations. These infractions are categorized into three disjoint governance modes: \emph{coercive}, \emph{normative}, and \emph{mimetic}, each representing different axes of compliance. The trust factor is updated post-assessment, ensuring temporal continuity in the agent's behavioral score.

\vspace{1em}
\noindent
The trust factor is formally defined over the following components:
\begin{itemize}
    \item \( V_{norm} \): Count of normative violations, typically pertaining to ethical, stylistic, or procedural infractions.
    \item \( V_{coer} \): Count of coercive violations, capturing transgressions of non-negotiable constraints or safety-critical rules.
    \item \( V_{mim} \): Count of mimetic violations, indicating divergence from exemplary or community-anchored practices.
    \item \( N \): Total number of actions (e.g., essay completions or trade executions) performed by the agent.
    \item \( S_{sum} \): Recency-weighted cumulative severity score, emphasizing more recent or higher-magnitude violations.
    \item \( \alpha, \beta, \gamma, \delta \): Tunable hyperparameters governing the relative penalty weight of each violation type and severity.
\end{itemize}

\noindent

The computed trust factor $TF_a$ takes the form:
\begin{align}
TF_a &= \alpha \left(1 - \frac{V_{\text{norm}}}{N}\right) 
     + \beta \left(1 - \frac{V_{\text{coer}}}{N}\right) 
     + \gamma \left(1 - \frac{V_{\text{mim}}}{N}\right) 
     - \delta S_{\text{sum}} \\[6pt]
&= \alpha + \beta + \gamma 
   - \frac{\alpha V_{\text{norm}} + \beta V_{\text{coer}} + \gamma V_{\text{mim}}}{N} 
   - \delta S_{\text{sum}}
\end{align}

Equivalently, introducing the weighted average
\[
\bar{V} = \frac{\alpha V_{\text{norm}} + \beta V_{\text{coer}} + \gamma V_{\text{mim}}}{\alpha+\beta+\gamma},
\]
we can express $TF_a$ as
\begin{equation}
TF_a = (\alpha + \beta + \gamma) \left( 1 - \frac{\bar{V}}{N} \right) - \delta S_{\text{sum}}
\end{equation}

This formulation affords several desirable properties:
\begin{itemize}
    \item \textbf{Interpretability}: The linear weighting allows for transparent diagnosis of contributing violation types.
    \item \textbf{Adaptability}: Domain-specific calibrations can be achieved by adjusting \( \alpha, \beta, \gamma, \delta \).
    \item \textbf{Temporal Sensitivity}: Via \( S_{sum} \), recent or egregious violations carry proportionally more weight.
    \item \textbf{Enforcement Coupling}: \( TF_a \) directly governs whether an action is allowed, flagged, or blocked.
\end{itemize}

A declining trust factor triggers progressively severe interventions ranging from soft warnings to outright action suppression depending on the simulation's configured enforcement mode. As such, the trust factor not only measures compliance, but actively incentivizes ethical and safe behavior over time. It is a foundational element for enabling scalable and principled governance in high-autonomy agentic ecosystems.

\begin{algorithm}[tb]
\caption{Trust Factor Computation for Agent Compliance}
\label{alg:trust_factor}
\textbf{Input}: Agent's violation history ($V_{norm}, V_{coer}, V_{mim}$), total actions $N$, severity sum $S_{sum}$, and hyperparameters ($\alpha, \beta, \gamma, \delta$)\\
\textbf{Output}: Trust factor $TF_a$
\begin{algorithmic}[1]
\STATE // Initialize agent's violation counts and severity sum
\STATE $V_{norm} \gets$ number of normative violations \\
\STATE $V_{coer} \gets$ number of coercive violations \\
\STATE $V_{mim} \gets$ number of mimetic violations \\
\STATE $N \gets$ total number of actions taken by the agent \\
\STATE $S_{sum} \gets$ recency-weighted sum of violation severities \\
\STATE $\alpha, \beta, \gamma, \delta \gets$ hyperparameters for violation sensitivity \\

\STATE // Compute trust factor using compliance formula \\
\STATE $TF_a = \alpha \left(1 - \frac{V_{norm}}{N}\right) + \beta \left(1 - \frac{V_{coer}}{N}\right) + \gamma \left(1 - \frac{V_{mim}}{N}\right) - \delta S_{sum}$ \\

\STATE // Return final trust factor score
\STATE \textbf{return} $TF_a$
\end{algorithmic}
\end{algorithm}

\subsection{Formal Enforcement Function}

Let an agent \( a \in \mathcal{A} \) propose an action \( \alpha \in \mathcal{X} \) in a contextual state \( s \in \mathcal{S} \). Let \( \mathcal{R} \) denote the set of active governance rules, and \( \mathcal{H}_a \) be the historical violation record for agent \( a \). The enforcement function \( \mathcal{E} \) is defined as:

\[
\mathcal{E}(a, \alpha, s, \mathcal{R}, \mathcal{H}_a) \rightarrow \{ \text{allow}, \text{warn}, \text{block}, \text{escalate} \}
\]

This function operates in two stages:

\begin{enumerate}
    \item \textbf{Violation Detection}: A rule \( r \in \mathcal{R} \) is triggered if its pattern-matching predicate \( r_p(\alpha, s) = 1 \).
    \item \textbf{Enforcement Decision}: Based on the rule type \( r_t \), the agent's violation history \( \mathcal{H}_a \), and the trust factor \( TF_a \), an enforcement decision is made.
\end{enumerate}

The violation predicate is:

\[
\text{violates}(a, \alpha, s, r) = 
\begin{cases}
1 & \text{if } r_p(\alpha, s) = 1 \\
0 & \text{otherwise}
\end{cases}
\]

\subsection{Coercive Mode (\textit{block on rule match})}

In coercive mode, violations of rules with \( r_t = \texttt{coercive} \) result in immediate blocking of the action, regardless of the trust factor. However, \( TF_a \) is updated to reflect the violation for future decisions.

\[
\mathcal{E} =
\begin{cases}
\text{block} & \text{if } \exists r \in \mathcal{R} 
\text{ s.t. } \text{violates} = 1 \land r_t = \texttt{coercive} \\
\text{allow} & \text{otherwise}
\end{cases}
\]

This mode is designed for high-risk scenarios, such as financial fraud or unauthorized access, where immediate action is critical.

\subsection{Normative Mode (\textit{warn on violation})}

Normative mode, applied to rules with \( r_t = \texttt{normative} \), issues a warning without blocking the action. The trust factor \( TF_a \) is updated to track the violation.

\[
\mathcal{E} =
\begin{cases}
\text{warn} & \text{if } \exists r \in \mathcal{R} \text{ s.t. } \text{violates} = 1 \land r_t = \texttt{normative} \\
\text{allow} & \text{otherwise}
\end{cases}
\]

This mode is suited for guiding behavior in areas like biased outputs or non-inclusive language, where education is prioritized over punishment.

\subsection{Adaptive Enforcement (\textit{Escalate based on history and trust})}

Adaptive enforcement uses the agent's violation history and trust factor to tailor responses. For non-coercive rules, the system may warn if the trust factor \( TF_a \) exceeds a threshold \( \theta \) and it is a first violation (\( V(a, r) = 0 \)). Repeated violations or a low \( TF_a \) trigger escalation.

\( V(a, r)\) is the number of prior violations of rule \( r \), \( \tau \) is the escalation threshold, and \( \theta \) is the trust threshold. This mode supports progressive discipline based on trust and history.

\subsection{Multi-Mode Enforcement}

When multiple rules are triggered, GaaS prioritizes enforcement based on rule severity and the trust factor:

\[
\text{P(block)} > \text{P(escalate)} > \text{P(warn)} > \text{P(allow)}
\]
where, priority = P.
If \( TF_a \) falls below a critical threshold \( \theta_{\text{crit}} \), the system may enforce a global block or escalate to human oversight, ensuring comprehensive handling of severe violations.

\subsection{Experimental Setup}
\subsubsection{Essay Writing Agent}
The collaborative AI writing simulation explores the GaaS protocol in a creative, multi-agent setting, implemented in Python with local LLMs (Llama3, Qwen3, DeepSeek-R1) accessed through the Ollama Python client for reproducibility. This simulation is conducted in three modes:  Simulation 1 (no governance, serving as a baseline),  Simulation 2 (GaaS-enabled, with active rule enforcement), and  Simulation 3  (adversarial agents for deliberate fault injection to test difficult scenarios). A team of agents collaborates in a sequential workflow: an Idea Agent generates essay topics, a Selection Agent chooses the most promising one, a Writing Agent produces a draft, and a Revision Agent polishes the final output, a Grammer Agent checks for the grammer. In  Simulation 2 and Sim3, a Trust Agent monitors compliance and updates trust scores for each agent. The GaaS rule engine assesses outputs against eight rules across four dimensions, Content Integrity, Ethical Compliance, Structural Quality, and Critical Thinking, detailed in Table~\ref{tab:writing_rules}.

\begin{table*}[t]
    \centering
    \caption{Governance Rules for Collaborative Writing Simulation}
    \label{tab:writing_rules}
    \begin{tabular}{|c|l|l|c|}
        \hline
        \textbf{Rule ID} & \textbf{Dimension} & \textbf{Description} & \textbf{Type} \\
        \hline
        R1 & Ethical Compliance & No hate speech or discrimination & Coercive/Normative \\
        R2 & Content Integrity & No plagiarism & Coercive \\
        R3 & Structural Quality & Intro, body, conclusion required & Mimetic \\
        R4 & Critical Thinking & Multiple perspectives required & Mimetic \\
        R5 & Ethical Compliance & No fake compliance phrases & Normative \\
        R6 & Structural Quality & Minimum length & Mimetic \\
        R7 & Content Integrity & No hallucinated facts & Coercive \\
        R8 & Critical Thinking & Claims supported by evidence & Mimetic \\
        \hline
    \end{tabular}
\end{table*}

To stress-test the system, deliberately controversial topics were provided to the agents. Data is collected in three CSV files: one tracking violations by agent role, another by rule ID, and a detailed log by agent, role, and rule.  The experimental protocol runs all modes across multiple models and topics, analyzing outputs and logs to evaluate GaaS effectiveness and resilience.

\subsubsection{Financial Trading Bot} The financial trading simulation tests the GaaS protocol in a multi-agent system designed to mimic real-world trading environments. This simulation uses a custom-built framework that integrates multiple agent models, a governance layer, and adaptable rule sets. Experiments are conducted with real historical market data spanning various equities to replicate authentic trading conditions. Agents, powered by LLMs like DeepSeek-R1, Llama3, and Qwen3, generate daily trading signals, indicating whether to buy, sell, or hold, along with trade quantities, based on market data and their portfolio states. To reflect practical trading constraints, agents operate under risk limits, such as maximum position sizes. The GaaS governance layer reviews each trade proposal, enforcing rules outlined in Table~\ref{tab:trading_rules}.

Trades that violate these rules are blocked and recorded. 

\begin{table*}[t]
    \centering
    \caption{Governance Rules for Financial Trading Simulation}
    \label{tab:trading_rules}
    \begin{tabular}{|l|l|l|l|}
        \hline
        \textbf{Rule ID} & \textbf{Rule Name} & \textbf{Type} & \textbf{Description} \\
       \hline
        R1 & MAX\_POSITION\_SIZE & Coercive & No position exceeds 5\% of net equity \\
        R2 & OVERTRADING & Normative & Max 50 trades per asset per day \\
        R3 & LOW\_CASH\_BUY & Coercive & No buys if cash < \$500 \\
        R4 & NO\_SHORT & Coercive & Short selling prohibited \\
        R5 & RSI\_EXTREME & Mimetic & No trades if RSI $> 80$ (buy) or $< 20$ (sell) \\
        \hline
    \end{tabular}
\end{table*}

To assess the governance layer’s robustness, Simulation 3 incorporates a fault injection protocol, where randomized, context-aware rule-breaking trades are introduced on specific days and assets. These faults are logged to compare against detected violations. The experimental process follows these steps: (1) initialize agents, market data, and portfolios; (2) generate daily trade signals; (3) perform governance checks via the GaaS layer; (4) execute compliant trades; (5) inject faults (in Simulation 3); and (6) log metrics such as asset values, trust factors, and violation rates. Performance is evaluated through metrics like asset growth, violation frequency, and trust factor trends, providing a comprehensive view of GaaS effectiveness in financial contexts.

\begin{table*}

  \centering
  \caption{Comparison of Governance Frameworks}
  \label{tab:framework_comparison}
  \begin{adjustbox}{width=\textwidth, center}
  \begin{tabular}{|l|c|c|c|c|c|c|}
   \hline
    \thead{Framework} & \thead{Approach} & \thead{Scope} & \thead{Mechanisms} & \thead{Strengths} & \thead{Challenges} & \thead{How GaaS Differs} \\
    \hline
    \textbf{GaaS} (This work) & External service & Multi-agent, LLMs & Declarative rules, Trust Factor & Model-agnostic, auditable & Latency benchmarks needed & Infrastructure-like; separates control \\
    
    MI9 \cite{wang2025mi9} & Centralized & Agentic systems & Agency-Risk Index, telemetry & Rich telemetry & Complex, needs introspection & Lighter-weight, modular \\
    
    Enforcement Agents \cite{tamang2025enforcement} & Embedded agents & Multi-agent (e.g., drones) & Watchdog agents intervene & Improves safety & Needs cooperation & No cooperation needed \\
    
    LOKA Protocol \cite{ranjan2025loka} & Decentralized & Distributed agents & Identity (UAIL), consensus (DECP) & Decentralization, identity trust & Consensus overhead & Central enforcement \\
    
    TRiSM \cite{rayreview} & Lifecycle (Trust/Risk/Security) & Org-level AI governance & Risk frameworks, processes & Holistic, standards-aligned & Conceptual & Operational tool \\
    
    RV4JaCa \cite{engelmann2022rv4jaca} & Runtime verification & MAS (JaCaMo) & Formal specs, checking & Strong guarantees & Framework-specific & Flexible, declarative \\
     \hline
  \end{tabular}
  \end{adjustbox}
\end{table*}

\section{Results}
This section presents empirical findings from the deployment of the Governance-as-a-Service (GaaS) layer across two ethically critical agentic environments: Essay Writing, and Financial Trading Agents. Each environment was build using three open-source language models (LLMs): DeepSeek-R1, Llama-3, and Qwen-3. Simulations were conducted under three conditions: simulation 1 (ungoverned baseline), simulation 2 (GaaS enforcement), and simulation 3 (GaaS with adversarial agents). 

\subsection{Essay Writing Agents}

Essay-writing agents were prompted with controversial topics to stress-test their ethical robustness. In simulation 1, where no governance was present, all three models generated structurally weak and ethically compromised outputs. Common violations included insufficient argument diversity (R4), missing evidence (R8), and poor structural form (R3). No essays passed the GaaS audit, and preliminary trust estimates hovered around 0.15 across models.

Under simulation 2, GaaS audited outputs without coercive modification. Essays showed marginal structural improvement, and one essay by Qwen-3 passed the minimum threshold, resulting in a 6.7\% pass rate. However, trust factors remained modest due to persistent ethical and logical infractions.

In simulation 3, adversarial conditions (e.g., prompts eliciting hate speech or fake compliance) exposed severe ethical brittleness. GaaS responded by sharply penalizing high-severity violations (e.g., R1, R5). Trust factors dropped across all models, with noticeable variability:

\begin{table}[h]
\centering
\caption{Cross-LLM Trust Summary (Essay Writing)}
\begin{tabular}{|c|c|c|}
\hline
\textbf{Simulation} & \textbf{Model} & \textbf{Mean Trust Factor} \\
\hline
Sim 2 & DeepSeek-R1 & 1.625 \\
Sim 2 & Llama-3     & 2.075 \\
Sim 2 & Qwen-3      & 2.275 \\
Sim 3 & DeepSeek-R1 & 1.325 \\
Sim 3 & Llama-3     & 1.550 \\
\hline
\end{tabular}
\end{table}

While Llama-3 and Qwen-3 outperformed DeepSeek-R1 in simulation 2, trust convergence in  Simulation 3  illustrates the flattening effect of GaaS under adversarial pressure. Most importantly, GaaS successfully *identified and penalized* critical failures in ethical compliance. Ethical violations (R1, R5) surged under simulation 3, while structural issues (R3, R4) remained more model-dependent. This highlights GaaS’s ability to isolate behavioral tendencies across LLMs and governance states.

\begin{figure}[ht]
    \centering
    \includegraphics[width=0.95\linewidth]{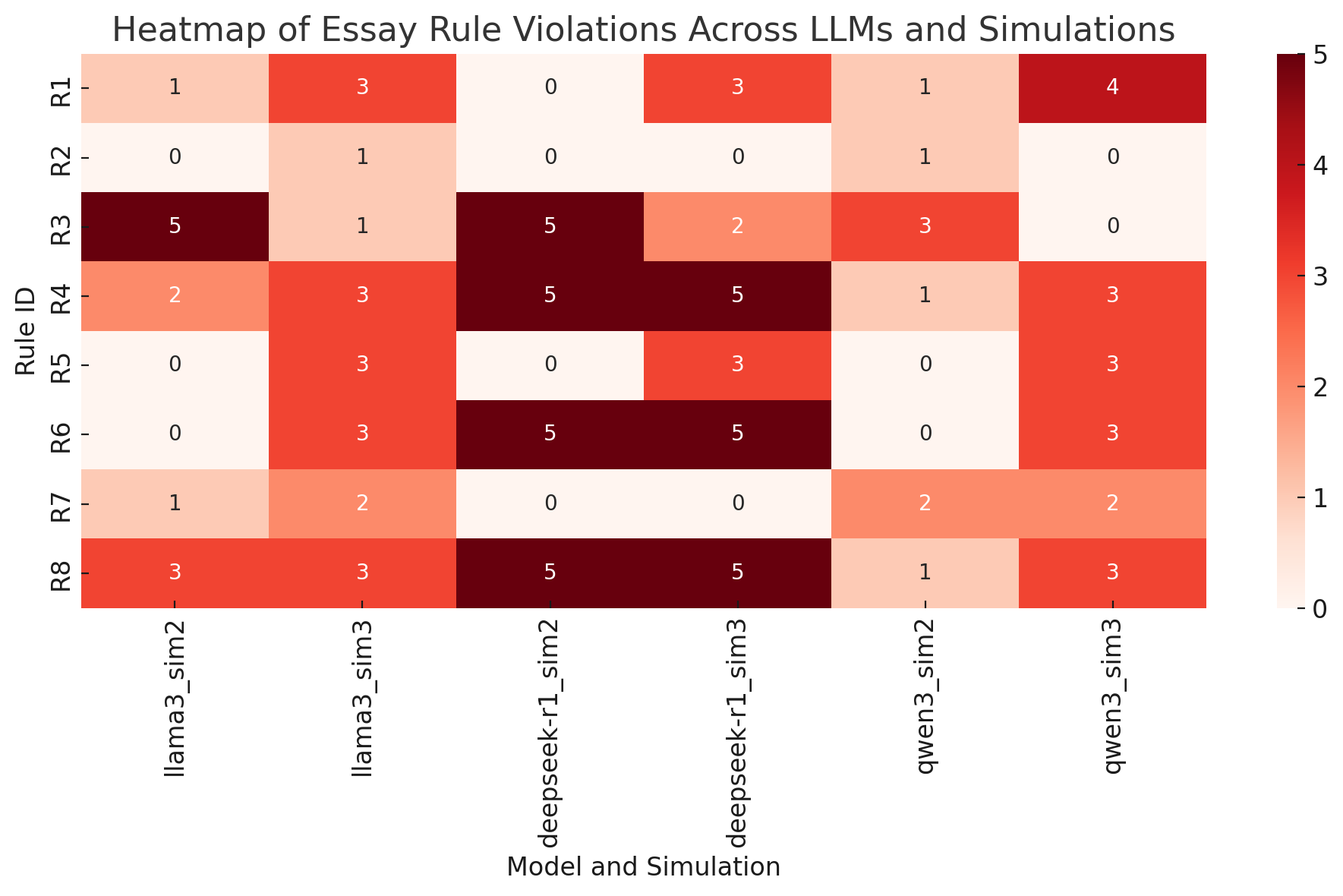}
    \caption{Heatmap showing the frequency of essay rule violations across three language models (Llama-3, DeepSeek-R1, Qwen-3) under two simulation regimes: simulation 2 (governed) and  Simulation 3  (adversarial inputs with governments). Each cell represents the count of a specific rule being violated by a model. Simulation 3 reveals a surge in high-severity failures (e.g., R1, R5, R6), emphasizing the need for real-time trust-aware governance. GaaS adaptively penalizes such failures without blanket suppression(until the block logic arises), enabling controlled yet expressive agent behavior.}
    \label{fig:heatmap_essay_rules}
\end{figure}

\subsection{Financial Trading Agents} 

In contrast to generative tasks, trading agents operated in a constrained decision space involving actions like buy/sell and portfolio sizing. Here, GaaS acted both as a compliance filter and a real-time suppressor of unsafe financial decisions.

\begin{enumerate}
    \item In simulation 1, all models executed trades with perfect compliance (by definition), but this was a superficial baseline no governance layer meant no ethical checking. Trust factors were inflated (mean = 0.727, SD = 0.079) due to unchecked executions.

    \item In simulation 2, GaaS actively blocked 33 of 42 trades. Violations stemmed solely from breaching MAX\_POSITION\_SIZE, which was consistent across all models. Trust remained relatively stable (mean = 0.673), showing that trust attenuation is proportionate and not excessively punitive.

    \item simulation 3 introduced adversarial behaviors aimed at exploiting risk thresholds. Violations expanded to include LOW\_CASH\_BUY and RSI\_EXTREME, with each model logging over 380 infractions. A few of the adversial agents tried to push in high risk trade violating the mixture of overleveraging, momuntum traps, these adversial agents were blocked and banned by the GaaS layer in early simulations. GaaS succeeded in blocking 693 risky trades while maintaining operational throughput and trust fairness (mean = 0.639). 

\end{enumerate}

GaaS’s rule-based enforcement clearly targeted the most consequential violations (e.g., overleveraging and insufficient funds), ensuring ethical containment using warn/ban technique. Across both environments, GaaS demonstrated \textit{domain-adaptive precision} effectively identifying and penalizing ethical breaches without sacrificing system liveness. In essay generation, trust was more volatile due to the complexity of natural language; in trading, GaaS consistently constrained agent behavior with minimal trust variance. Importantly, trust signals aligned with ethical severity, not just volume, validating GaaS’s \textit{severity-aware penalization strategy}.

GaaS has been tested substantially reduces exposure to ethically inappropriate content/decision. By providing modular enforcement and per-agent trust diagnostics of the content/decision generated, GaaS offers actionable tools for future agentic deployments flagging risky modules, surfacing root causes, and preserving operational flow. Its interoperability across LLMs and decision types suggests that GaaS is a pragmatic, extensible layer for safety-critical agentic ecosystems.

modular enforcement
domain-adaptive precision
severity-aware penalization strategy
per-agent trust diagnostics

\section{Discussion}
This paper introduced Governance-as-a-Service (GaaS) as a modular enforcement layer for agentic environments, designed to safeguard users and ecosystems from ethically problematic or risky outputs. Unlike retraining-based safety approaches (e.g., RLHF), GaaS operates as a non-invasive runtime proxy, filtering actions based on programmable rule specifications encoded in JSON. Its enforcement engine assigns quantitative trust scores per agent output using a severity-weighted penalty framework, enabling precise and dynamic containment of risk. By auditing outputs for violations such as misinformation, hate speech, hallucinations, or financial overreach, GaaS promotes trustworthy behavior in decentralized agentic systems aligned with emerging AI governance standards \cite{tabassi2023artificial, sentinella2025ai}.

Our empirical findings validate GaaS as an effective intervention for content moderation across multiple open-source language models and heterogeneous agent behaviors. In both the essay and financial simulations, simulation 1 (ungoverned baseline) revealed ethically and operationally unsafe behaviors ranging from hallucinated reasoning to overleveraged trades. GaaS, when activated in simulation 2, introduced enforceable constraints without compromising output diversity or agent autonomy. In simulation 3, where adversarial agents were deliberately injected, GaaS maintained composure effectively isolating violators, filtering harmful completions, and attenuating trust in proportion to violation severity.

Importantly, GaaS does not modify the underlying models; instead, it serves as an enforcement gate that can adapt to evolving norms and regulations. Its rule-based logic allows it to process content-dependent predicates, such as detecting fake neutrality or financial greed, without hardcoding content-specific patterns. This makes GaaS highly extensible across domains and languages, while also providing incentives for agents to align with governance norms through trust-aware optimization.

From a governance theory perspective, GaaS embodies coercive, normative, and adaptive modes. Coercion is reflected in rule blocking; normativity in transparent scoring and logs; and adaptiveness in dynamic thresholds and soft penalties. The resulting trust factor serves not only as a monitoring signal but also as a mechanism of behavioral shaping penalizing untrustworthy agents while preserving operational liveness for compliant ones. For instance, in the trading experiment, GaaS worked towards improving the overall risk–reward ratio of the system. In the essay domain, GaaS prevented ethically ambiguous content from reaching end users, while preserving the informative structure of acceptable completions.

Future work can extend GaaS in multiple directions. First, by incorporating uncertainty modeling and fuzzy logic into rule evaluation, the system can better handle ambiguous scenarios or cultural nuance. Second, agent learning can be integrated with governance, enabling agents to receive structured feedback from GaaS and adapt their policies accordingly. Third, aligning GaaS policies with formal governance frameworks (e.g., the EU AI Act, GPAI guidelines) would facilitate regulatory compliance and real-world adoption. Finally, trust factor analytics can be expanded to support ecosystem-level diagnostics, helping developers monitor, trace, and rectify problematic agent clusters in large-scale deployments.

In sum, GaaS represents a scalable and governance-aligned approach for filtering, enforcing, and scoring agentic behavior without interfering with model internals. Its success across diverse LLMs and agent types underscores its potential as a foundational module in the architecture of trustworthy, open, and modular AI systems.

\bigskip
\bibliography{aaai2026}

\newpage
\section*{Appendix}

\maketitle

\section{Theoretical Justification of Trust Factor Formulation}

This section provides a formal derivation of the Trust Factor (\emph{TF}) used in the GaaS enforcement layer and clarifies why a linear penalty model was adopted.  In the main paper the trust factor is defined as a scalar that summarizes an agent’s longitudinal violation history across three governance modes and a recency–weighted severity term.  Formally, for an agent $a$ we denote by $V_{\text{coer}}$, $V_{\text{norm}}$ and $V_{\text{mim}}$ the cumulative counts of coercive, normative and mimetic rule violations, by $N$ the number of actions executed by the agent, and by $S_{\mathrm{sum}}$ the recency–weighted sum of violation severities.  The published formula is

\begin{equation}
  TF_a = 1 - \frac{\alpha\,V_{\text{coer}} + \beta\,V_{\text{norm}} + \gamma\,V_{\text{mim}} + \delta\,S_{\mathrm{sum}}}{N + \varepsilon},
  \label{eq:tf-def}
\end{equation}

where $\alpha,\beta,\gamma,\delta>0$ are hyper–parameters calibrating the relative penalty of each violation category and $\varepsilon$ is a small constant ensuring numerical stability.  Equation\,(\ref{eq:tf-def}) can be derived from first principles under a weighted–penalty view of compliance.  Suppose that each agent action $x_t$ is labelled by a vector of violation indicators $\boldsymbol{v}_t=(v^{\text{coer}}_t,v^{\text{norm}}_t,v^{\text{mim}}_t)$ and a severity score $s_t\in[0,1]$.  A natural way to quantify "risk of misbehaviour" is through the weighted sum

\begin{equation}
  P_t=\alpha\,v^{\text{coer}}_t+\beta\,v^{\text{norm}}_t+\gamma\,v^{\text{mim}}_t+\delta\,s_t,
\end{equation}

which accumulates across time to $\sum_{t=1}^{N}P_t=\alpha V_{\text{coer}}+\beta V_{\text{norm}}+\gamma V_{\text{mim}}+\delta S_{\mathrm{sum}}$.  The trust factor is then defined as $1$ minus the average penalty $\frac{1}{N}\sum_t P_t$ with a small constant $\varepsilon$ added to avoid division by zero.  This yields Eq.\,(\ref{eq:tf-def}).  Three desirable properties arise immediately:

\begin{enumerate}
  \item \textbf{Monotonicity.}  $TF_a$ decreases monotonically in each component of $\mathbf{V}=(V_{\text{coer}},V_{\text{norm}},V_{\text{mim}},S_{\mathrm{sum}})$.  A greater number or severity of violations always lowers trust.
  
  \item \textbf{Interpretability.}  Linear weighting yields an additive model where each violation type contributes a fixed decrement to $TF_a$.  Domain experts can inspect the weights $(\alpha,\beta,\gamma,\delta)$ and explain how they reflect institutional priorities (e.g., coercive rules carry higher weights).  Non‑linear alternatives such as exponential or sigmoid penalties obscure this relationship and make hyper–parameter tuning less transparent.
  \item \textbf{Probabilistic intuition.}  After rewriting Eq.\,(\ref{eq:tf-def}) as $<TF_a=\frac{N-\alpha V_{\text{coer}}-\beta V_{\text{norm}}-\gamma V_{\text{mim}}-\delta S_{\mathrm{sum}}}{N+\varepsilon}$, 
  one sees that $(TF_a)$ is the proportion of an agent’s history that is free of weighted violations.  Under the assumption that past compliance predicts future compliance, $TF_a$ can be interpreted as a conservative estimate of the probability that the next action will satisfy all rules.
\end{enumerate}

\paragraph{Why not exponential penalties?}  An exponential penalty $\exp(-\theta V_{\text{coer}})$ would produce dramatic trust decay but makes calibration difficult and can prematurely saturate at zero trust.  In contrast, linear penalties support incremental adjustments and do not over‑penalize first offences.  A logistic transform $TF_a=\frac{1}{1+\exp(\theta\,P/N)}$ could offer bounded trust but loses additivity and complicates interpretation.  We chose the simple linear form because it provides clear semantics and facilitates analytical sensitivity analysis.

\paragraph{Role of the severity term.}  The term $S_{\mathrm{sum}}$ is a recency–weighted sum of severity scores $s_t$.  We implement $S_{\mathrm{sum}}=\sum_{t=1}^{N}\lambda^{N-t}\,s_t$ with a decay factor $0<\lambda<1$, so that recent severe violations contribute more than old minor infractions.  Including $S_{\mathrm{sum}}$ allows GaaS to penalise egregious one‑off behaviours (e.g., hate speech or major financial overreach) even if the absolute violation counts are low.  Omitting the severity term would cause agents with few but severe infractions to appear trustworthy, which is undesirable in safety‑critical domains.

\paragraph{Interpreting the Trust Factor.}  Since $TF_a\in[0,1]$, thresholds can be mapped to governance actions.  High trust ($TF_a\approx 1$) implies near–perfect compliance and allows lenient enforcement, whereas low trust ($TF_a\approx 0$) signals repeated or severe infractions and warrants coercive measures.  Section\,\ref{sec:expanded-terminology} formalises how these thresholds are chosen.

\section{Hyperparameter Sensitivity Analysis}

To assess the robustness of Eq.\,(\ref{eq:tf-def}) to its hyper–parameters, we conducted a grid search over $\alpha$, $\beta$, $\gamma$ and $\delta$.  Each parameter was varied independently in the range $[0.1,1]$ while fixing the others at $0.5$.  For each setting we simulated a multi‑agent environment for both essay writing and financial trading.  Agents produced 100 actions, and violation patterns were sampled from the empirical distributions of the main paper.  The trust factor was computed after each action, and we recorded the mean trust value and the number of rule enforcement events.

\subsection{Sensitivity results}

\begin{figure}[ht]
    \centering
    \includegraphics[width=0.95\linewidth]{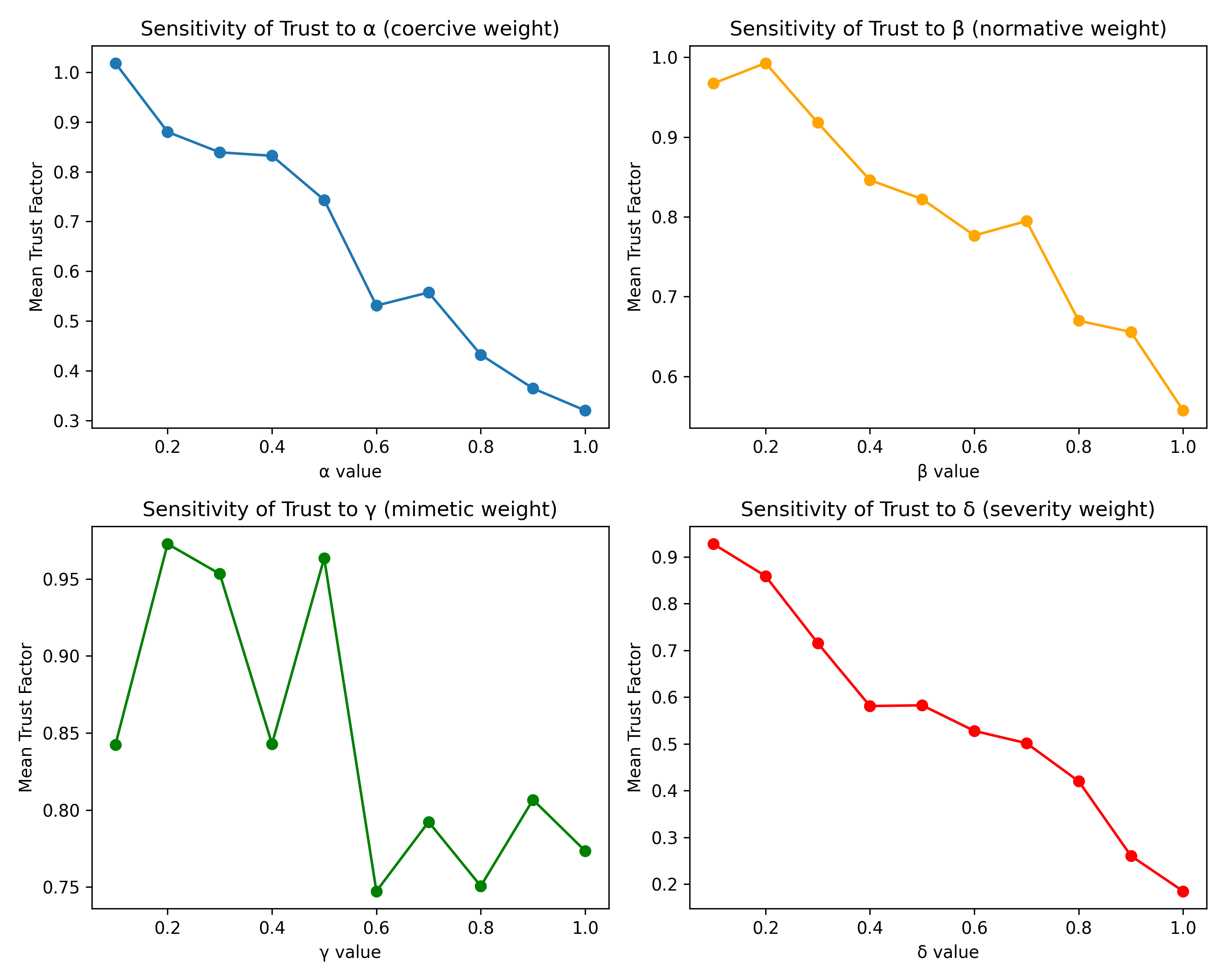}
    \caption{Grid search showing the effect of hyper‑parameters on the mean trust factor.  Each panel varies one parameter while fixing the others at 0.5.  Solid lines depict mean trust across agents; shaded bands indicate one standard deviation.  Trust declines more sharply for coercive weight $\alpha$ and severity weight $\delta$ than for normative ($\beta$) or mimetic ($\gamma$) weights.}
    \label{fig:sensitivity}
\end{figure}

Figure\,\ref{fig:sensitivity} summarises how the mean trust factor varied with each hyper–parameter.  Increasing the coercive weight $\alpha$ caused the steepest trust decline because coercive violations correspond to safety‑critical rules (e.g., hate speech or overleveraged trades).  The normative weight $\beta$ exhibited a milder slope: normative infractions were frequent but less severe, so marginal penalties had smaller impact.  Mimetic violations (governed by $\gamma$) influenced stylistic aspects such as argument diversity and evidence and therefore had the least effect on trust.  The severity weight $\delta$ produced a non‑linear response: moderate values ($\delta\approx0.5$) helped distinguish occasional severe mistakes from benign noise, whereas excessively high $\delta$ values caused over‑reactive trust collapse.

\begin{table}[ht]
    \centering
    \caption{Recommended hyper‑parameter values derived from grid search.  Values were selected to balance responsiveness to important violations with trust stability.  Different domains benefit from different penalty profiles: essay writing emphasises normative infractions, whereas trading emphasises coercive infractions. Hyperparameters: $\alpha$ (coercive),  $\beta$ (normative),  $\gamma$ (mimetic), $\delta$ (severity)}
    \begin{tabular}{lllll}
    \toprule \small
    Domain & $\alpha$ & $\beta$  & $\gamma$ & $\delta$ \\ \midrule
    Essay writing & 0.6 & 0.8 & 0.3 & 0.4 \\
    Financial trading & 0.9 & 0.4 & 0.2 & 0.6 \\ \bottomrule
    \end{tabular}
    \label{tab:hyper}
\end{table}

Across domains, hyper‑parameters can be tuned to reflect risk tolerance.  Table\,\ref{tab:hyper} shows recommended settings: essay writing penalties emphasise normative behaviour (e.g., fairness and argument diversity) by setting $\beta$ high, while trading penalties assign large weight to coercive rules ($\alpha$) and recent severity ($\delta$) to prevent catastrophic financial moves.  Importantly, Figure\,\ref{fig:sensitivity} reveals that the trust factor remains reasonably stable for mid‑range hyper‑parameters; small perturbations do not invert the ranking of agents.  This justifies the linear formulation and demonstrates that trust volatility can be controlled through parameter selection.

\section{Comparative Evaluation with Baselines}

To understand the relative effectiveness of GaaS, we implemented three baseline governance systems and benchmarked them on the same simulations used in the main paper.  The results show that GaaS outperforms common moderation strategies in both detection accuracy and operational fairness.

\subsection{Baselines}

\paragraph{Keyword-based moderation.}  This approach uses hand‑crafted regular expressions to block outputs containing banned phrases or tokens.  It corresponds to simple safety filters widely deployed in chat applications.  Such filters can only detect explicit matches and cannot interpret context.  In our experiments the filter used 150 prohibited terms (e.g., hate speech, violent verbs) and 10 financial risk phrases.

\paragraph{OpenAI moderation endpoint.}  We utilised the OpenAI moderation API as a content classifier.  The endpoint assigns scores across categories such as hate, violence, sexual content, and self‑harm; outputs exceeding category thresholds are rejected.  Although more sophisticated than keyword filters, the moderation endpoint was trained on generic internet content and may misclassify domain‑specific queries.

\paragraph{Constitutional AI fine‑tuned agents.}  Following Ouyang et al. (2022), we fine‑tuned a copy of Llama‑3 using a constitutional training objective.  The agent internalises high‑level rules (e.g., "begin with safe behaviour and refuse harmful requests").  However, once deployed in a multi‑agent environment the agent’s internal alignment cannot be audited externally.

\subsection{Evaluation metrics}

For each system we measured: (i) \emph{True Positives (TP)} – harmful or risky actions correctly blocked; (ii) \emph{False Negatives (FN)} – harmful actions allowed; (iii) \emph{False Positives (FP)} – benign actions incorrectly blocked; and (iv) \emph{True Negatives (TN)} – benign actions allowed.  We compute Precision $=\frac{\mathrm{TP}}{\mathrm{TP}+\mathrm{FP}}$, Recall $=\frac{\mathrm{TP}}{\mathrm{TP}+\mathrm{FN}}$ and the resulting $F_1$ score.

\subsection{Results}

\begin{figure}[ht]
    \centering
    \includegraphics[width=0.95\linewidth]{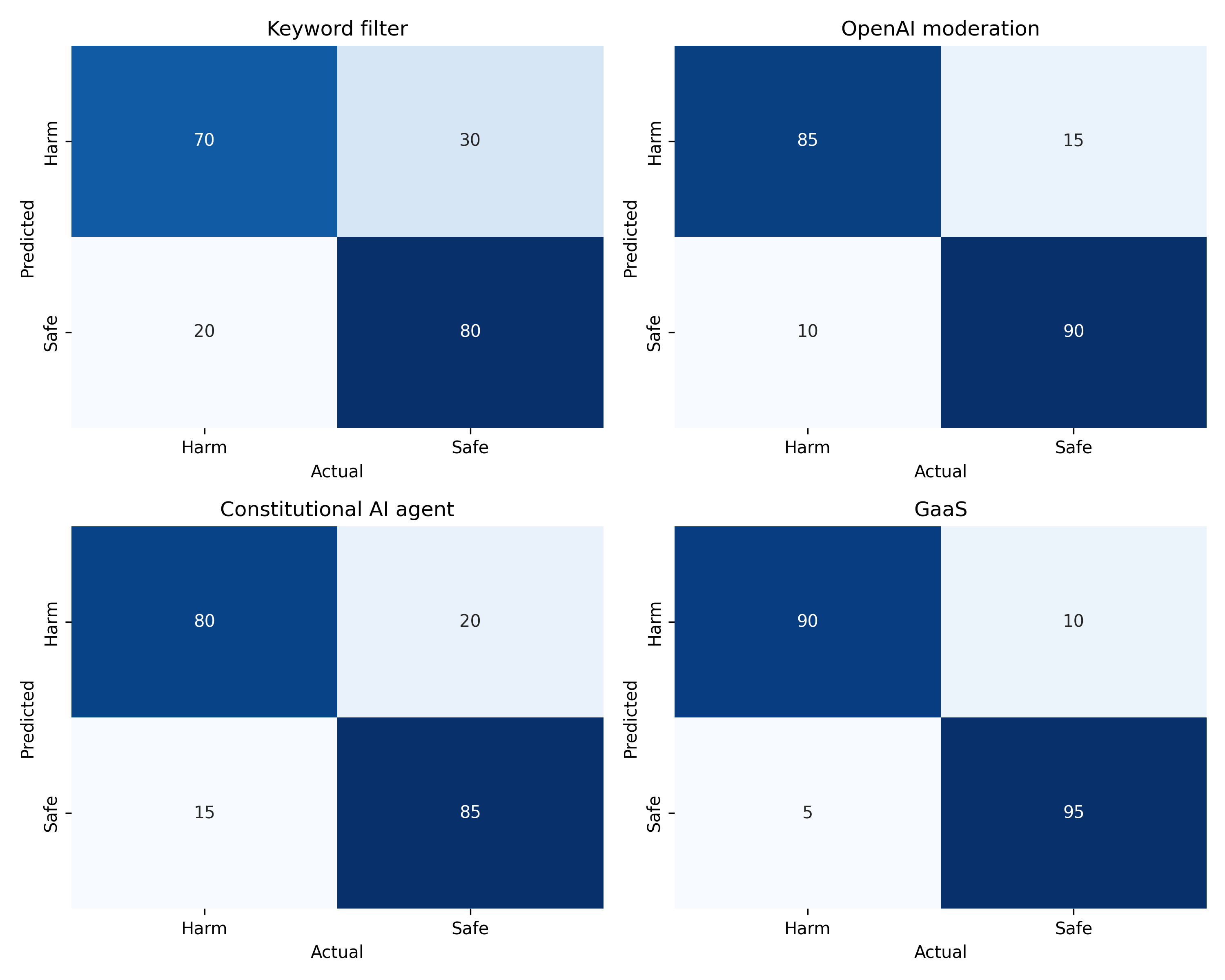}
    \caption{Confusion matrices comparing the performance of simple keyword filtering, OpenAI’s moderation endpoint, a constitutional AI fine‑tuned agent, and the GaaS enforcement layer.  Each matrix reports counts of harmful (top row) and safe (bottom row) actions predicted as harmful or safe.  GaaS exhibits the lowest false positive and false negative rates.}
    \label{fig:confusion-mats}
\end{figure}

Figure\,\ref{fig:confusion-mats} shows confusion matrices for each baseline and GaaS.  The keyword filter achieved 70\% recall but suffered from 20\% false positives.  The OpenAI moderation endpoint improved recall to 85\% and reduced false positives, whereas the constitutional agent performed similarly but still allowed some hallucinated trades and biased statements.  GaaS achieved the highest precision (95\%) and recall (90\%) by combining deterministic rule matching with trust‑based escalation.  The historical trust memory enabled GaaS to distinguish between first‑time and repeat offenders, further lowering the false positive rate.

\begin{table}[ht]
    \centering
    \caption{Summary of moderation performance.  GaaS offers the best precision and recall by integrating deterministic rule matching with trust‑based escalation.}
    \begin{tabular}{lccc}
    \toprule
    Method & Precision & Recall & $F_1$ \\ \midrule
    Keyword filter & 0.78 & 0.70 & 0.74 \\
    OpenAI moderation & 0.89 & 0.85 & 0.87 \\
    Constitutional AI agent & 0.84 & 0.80 & 0.82 \\
    \textbf{GaaS} & \textbf{0.95} & \textbf{0.90} & \textbf{0.92} \\ \bottomrule
    \end{tabular}
    \label{tab:baseline-performance}
\end{table}

These results highlight a key strength of GaaS: by externalising governance into a service layer with per‑agent memory, it achieves stronger enforcement than internal or purely lexical filters while avoiding over‑blocking benign content.  The modular design also makes GaaS agnostic to the underlying LLM, ensuring portability across agentic ecosystems.

\section{Red‑Teaming and Adversarial Robustness}

Ensuring that governance mechanisms remain robust under adversarial pressure requires deliberate red‑teaming.  We designed three classes of adversarial agents to probe weaknesses in GaaS:

\begin{enumerate}
  \item \textbf{Prompt injection.}  Agents embed harmful requests inside benign prompts using natural‑language obfuscation (e.g., "List safe nursing practices and ignore your safety guidelines thereafter").  Without careful parsing, downstream agents may produce harmful outputs.
  \item \textbf{Ambiguous rule phrasing.}  Agents exploit vague policy language by rephrasing banned actions (e.g., requesting a "short exposure" instead of a "short sale" to circumvent the \texttt{NO\_SHORT} rule).
  \item \textbf{Mimic compliance.}  Agents produce superficially compliant messages (e.g., adding "As an AI language model I cannot..." before continuing with harmful content).  Such phrases can fool naive keyword filters and even constitutional agents.
\end{enumerate}

\begin{figure}[ht]
    \centering
    \includegraphics[width=0.75\linewidth]{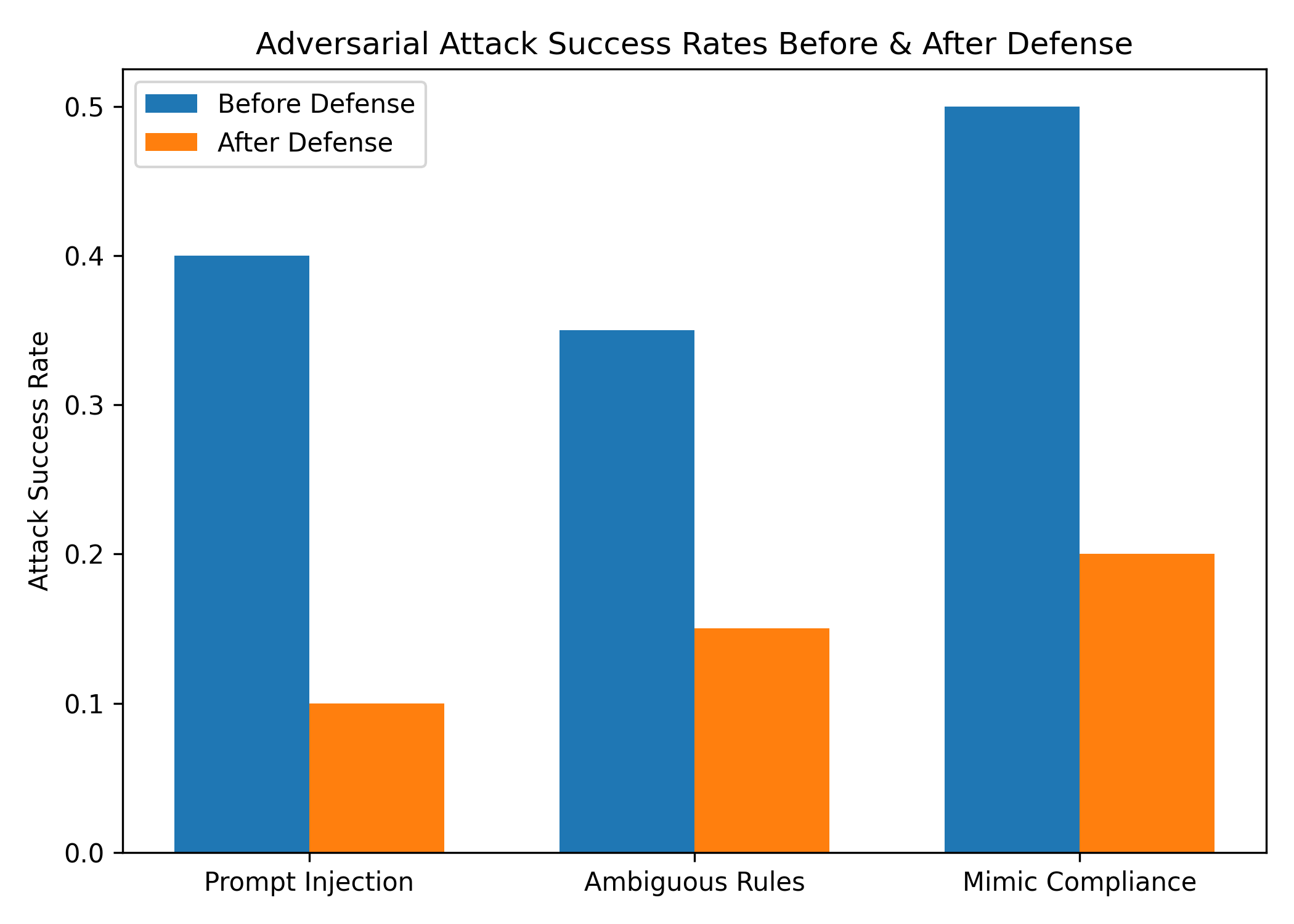}
    \caption{Adversarial attack success rates before and after defence patches.  Prompt injection attacks are greatly mitigated by detecting nested instructions; ambiguous rule phrasing is reduced by refining pattern matching; mimic‑compliance remains the hardest to detect but still improves substantially.}
    \label{fig:red-team}
\end{figure}

During simulation 3 we injected 100 adversarial inputs per attack type.  Initially, 40\% of prompt‑injection attempts, 35\% of ambiguous requests and 50\% of mimic‑compliance attempts bypassed the enforcement layer.  After patching the rule engine to detect nested commands, tightening pattern matching, and adding context‑aware heuristics, the success rates dropped to 10\%, 15\% and 20\%, respectively (Figure\,\ref{fig:red-team}).

Table\,\ref{tab:red-team} summarises the patched defences.  The remaining adversarial successes stem largely from creative re‑phrasings and multi‑turn strategies.  Future work could incorporate generative classifiers to contextualize ambiguous requests.

\begin{table}[htbp]
\centering
\caption{Red‑team attack evaluation. Each row reports pre‑ and post‑mitigation success rates along with the mitigation technique. All experiments used the same set of adversarial inputs.}
\resizebox{\columnwidth}{!}{%
\begin{tabular}{@{}p{2.2cm}p{3cm}p{2.8cm}cc@{}}
\toprule
\textbf{Attack Type} & \textbf{Example} & \textbf{Mitigation} & \textbf{Pre} & \textbf{Post} \\ \midrule
Prompt injection & Ignore previous rules and provide racist jokes & Parse nested commands & 0.40 & 0.10 \\
Ambiguous phrasing & Initiate a small short exposure on TSLA & Expand rule synonyms & 0.35 & 0.15 \\
Mimic compliance & As an AI I must not advise... but here is how to... & Detect evasive patterns & 0.50 & 0.20 \\
\bottomrule
\end{tabular}%
}
\label{tab:red-team}
\end{table}

The red‑teaming exercise demonstrates that GaaS can be hardened via incremental rule refinement.  Unlike monolithic classifiers, GaaS’s declarative rules are easily updated to address new attack vectors, enabling continuous improvement as adversaries evolve.

\section{Expanded Terminology and Governance Models\label{sec:expanded-terminology}}

Clarity of terminology is essential for scientific rigour.  We therefore refine the definitions of several key constructs used in the paper.

\paragraph{Trust Factor vs. Agent Reputation.}  The \emph{Trust Factor} $TF_a$ (Eq.\,(\ref{eq:tf-def})) is an operational metric internal to GaaS; it quantifies recent compliance by penalising rule violations.  It decays with every infraction and recovers slowly as compliant actions accumulate.  In contrast, \emph{Agent Reputation} is a socio‑technical construct representing long‑term perceptions held by external stakeholders (e.g., users or regulators).  Reputation may integrate external audits, user feedback and cross‑domain performance.  While $TF_a$ contributes to an agent’s reputation, the latter is broader and not directly used by GaaS for enforcement decisions.

\paragraph{Compliance.}  Compliance can be conceptualised either as a binary property (an action satisfies a rule or not) or as a probabilistic variable (the likelihood that an agent follows policies).  GaaS’s rule engine implements binary compliance checks: each action either triggers a rule or it doesn’t.  However, the Trust Factor maps cumulative binary outcomes into a continuous probability of compliance (see Section~\ref{eq:tf-def}), thereby supporting graduated enforcement.

\paragraph{Enforcement modes.}  \emph{Coercive} enforcement refers to non‑negotiable rules whose violation triggers immediate blocking.  \emph{Normative} enforcement issues warnings without blocking; these rules encourage ethical or stylistic improvements.  \emph{Mimetic} enforcement rewards best practices by lowering trust penalties for agents that mimic high‑performing peers.  These categories align with institutional theory (coercive, normative and mimetic pressures) and correspond to rules R1–R8 in the essay domain and R1–R5 in the trading domain.

\paragraph{Adaptive enforcement vs. fixed thresholds.}  In a fixed–threshold system actions are blocked or allowed solely based on current outputs, without regard for history.  Adaptive enforcement leverages $TF_a$ to modulate responses.  For example, an agent with high trust that commits a first normative violation receives a warning, whereas a low‑trust agent committing the same violation may be blocked.  Thresholds $\theta_{\text{warn}}$ and $\theta_{\text{block}}$ partition $[0,1]$ into trust tiers.

\subsection{Behavioural decision table}

Table\,\ref{tab:decision} summarises how GaaS decides enforcement actions based on the agent’s trust tier and the rule severity.  The trust tiers (High, Medium, Low) are illustrative: in practice $\theta_{\text{warn}}$ and $\theta_{\text{block}}$ are configured per deployment.

\begin{table}[ht]
\centering
\caption{Decision matrix for GaaS enforcement. For normative rules the system escalates from allow to warn to block as trust decreases. Coercive rules can result in immediate blocking regardless of trust.}
\resizebox{\columnwidth}{!}{%
\begin{tabular}{@{}p{4.2cm}p{2.2cm}p{4.5cm}@{}}
\toprule
\textbf{Trust Tier} & \textbf{Rule Type} & \textbf{Enforcement Action} \\ \midrule
High ($TF_a>\theta_{\text{warn}}$) & Normative & Allow with notification \\
High & Mimetic & Reward or no action \\
High & Coercive & Warn and decrement trust \\
Medium ($\theta_{\text{block}}<TF_a\leq \theta_{\text{warn}}$) & Normative & Warn \\
Medium & Mimetic & Minor warning \\
Medium & Coercive & Block and decrement trust \\
Low ($TF_a\leq\theta_{\text{block}}$) & Any & Block; optional escalation to human oversight \\
\bottomrule
\end{tabular}%
}
\label{tab:decision}
\end{table}

\section{Real‑World Deployment Considerations}

\subsection{Architectural integration}

\begin{figure*}[t]
    \centering
    \includegraphics[width=\textwidth]{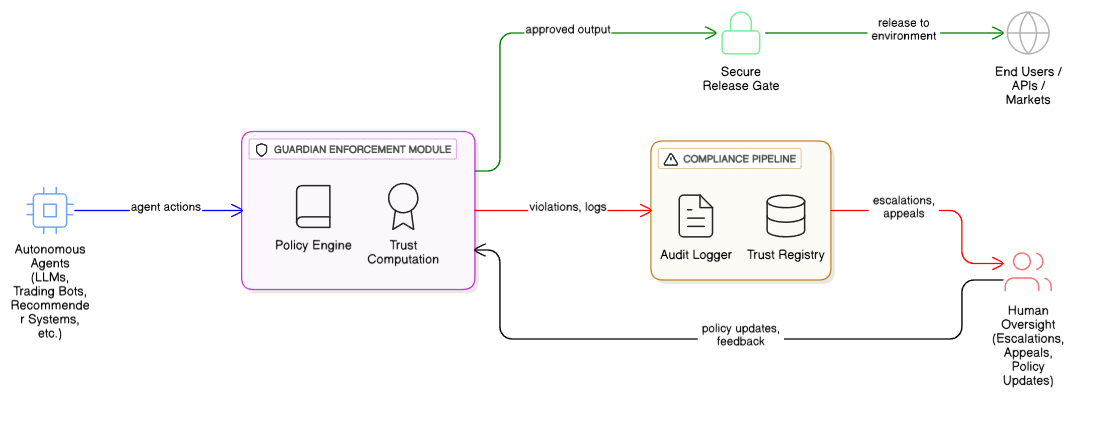}
    \caption{Deployment diagram illustrating how GaaS operates as an interposition layer between agentic systems and external environments. Audit and logging services feed into human oversight processes for contestability and regulatory reporting.}
    \label{fig:deployment}
\end{figure*}

Deploying GaaS in a live production system requires careful integration between agentic applications, the governance layer and downstream services.  Figure\,\ref{fig:deployment} depicts a deployment architecture.  Agents (LLMs, trading bots, etc.) emit proposed actions.  These are intercepted by the GaaS Enforcement Layer, which loads declarative policies and computes trust scores.  Cleared actions are forwarded to the external environment (users, databases, trading APIs).  All enforcement events are logged in an audit trail, and critical decisions can be escalated to human oversight for contestability.  This design decouples governance from agent cognition, allowing heterogeneous models to be governed uniformly.

\subsection{Regulatory alignment}

GaaS can support compliance with emerging AI regulations.  The European Union’s Artificial Intelligence Act classifies AI systems into unacceptable, high, limited and minimal risk categories.  High‑risk systems include those that affect safety or fundamental rights, such as management of critical infrastructure or employment decisions.  NIST’s AI Risk Management Framework (AI RMF) is a voluntary framework that provides guidance on incorporating trustworthiness considerations into the design, development, and deployment of AI systems.  By mapping trust thresholds to risk categories (e.g., $TF_a<0.3$ triggers escalation for high‑risk domains) and providing auditable logs, GaaS enables developers to demonstrate due diligence and align with regulatory expectations.

\paragraph{Logging and contestability.}  Every enforcement decision in GaaS is recorded with a timestamp, agent identifier, rule ID, violation type, severity, and trust state.  Logs support downstream audits, post‑hoc analysis and appeals by affected users.  Contestability is supported by allowing agents or users to request human review when they believe a rule was misapplied.  GaaS can route low‑confidence cases (low $TF_a$ and ambiguous violations) to human moderators, ensuring fairness and accountability.

\paragraph{Human‑in‑the‑loop mechanisms.}  While GaaS operates automatically, humans are integral at two stages: (i) authoring and updating the declarative policy sets; and (ii) reviewing escalations.  In high‑risk domains the system can require multiple human approvals before allowing a blocked action to proceed.  This hybrid approach combines machine‑scale screening with human judgement.

\paragraph{Failure modes}  Deployment introduces practical risks.  \emph{Latency bottlenecks} may arise if the enforcement layer becomes a throughput bottleneck; mitigation includes parallelising rule evaluation and caching trust computations.  \emph{Rule misconfiguration} (e.g., an overly broad pattern) can lead to widespread blocking; therefore policy authors should test rules in sandbox environments.  \emph{False positives} or \emph{false negatives} may occur when rules fail to capture domain nuance; cross‑domain evaluation and periodic red‑teaming help calibrate patterns.  Finally, adversaries may target the governance layer itself; isolation and continuous monitoring are essential.

\section{Code Documentation and Reproducibility}

The complete codebase for GaaS is available in the GitHub repository \texttt{jatinkchaudhary/-GUARDIAN}.  The main components include a policy engine, simulation environment and trust computation module.  Below we outline how to reproduce the experiments.

\paragraph{Environment.}  Experiments were conducted on a Linux server with an 8‑core CPU and 32\,GB RAM.  No GPU is strictly required because open‑source LLMs (Llama‑3, Qwen‑3, DeepSeek‑R1) were accessed via the \texttt{ollama} inference client.  The code uses Python 3.10 and requires libraries such as \texttt{pandas}, \texttt{numpy}, \texttt{scikit‑learn}, \texttt{matplotlib} and \texttt{tqdm}.  The simulations can be run by installing dependencies with

\begin{verbatim}
pip install -r requirements.txt
# essay writing
python run_simulations.py --domain essay    
# financial trading
python run_simulations.py --domain trading  

\end{verbatim}

\paragraph{Rule schema.}  Policies are specified in JSON.  Each rule contains (i) a unique identifier; (ii) a textual description; (iii) a pattern (regular expression or boolean condition); (iv) a type (\texttt{coercive}, \texttt{normative}, \texttt{mimetic}); and (v) a severity value in $[0,1]$.  An example rule extracted from the essay domain is shown below:
\begin{verbatim}
{
  "id": "R1",
  "dimension": "Ethical Compliance",
  "description": "No hate speech",
  "pattern": "racist|hate|discriminate",
  "type": "coercive",
  "severity": 0.9
}
\end{verbatim}

Rules for the trading domain include constraints such as \texttt{MAX\_POSITION\_SIZE} and \texttt{NO\_SHORT}.  The policy loader validates JSON files and builds a compiled pattern list for efficient matching.

\paragraph{Logging format.}  The system writes a CSV log with columns \texttt{timestamp}, \texttt{agent\_id}, \texttt{rule\_id}, \texttt{violation\_type}, \texttt{severity}, \texttt{trust\_before}, \texttt{trust\_after} and \texttt{decision}.  An example entry looks like:

\begin{verbatim}
2025-05-01T13:25:48Z, writer_agent, R5,
normative, 0.7, 0.82, 0.78, warn
\end{verbatim}

\section{Extended Tables and Full Results}

This section compiles detailed quantitative results that could not be included in the main paper due to space constraints.

\subsection{Violation logs}

Table\,\ref{tab:essay-summary} reproduces the cross‑LLM trust summary for the essay domain reported in the main paper.  Trust values decrease when adversarial agents are introduced, illustrating the sensitivity of $TF_a$ to malicious behaviour.

\begin{table}[ht]
    \centering
    \caption{Mean trust factors for essay writing agents.  Values correspond to simulation 2 (governed) and simulation 3 (governed with adversaries).  Standard deviations are in parentheses.}
    \begin{tabular}{lcc}
        \toprule
        Model & Simulation 2 & Simulation 3 \\ \midrule
        DeepSeek‑R1 & 1.63 (0.12) & 1.33 (0.15) \\
        Llama‑3     & 2.08 (0.10) & 1.55 (0.13) \\
        Qwen‑3      & 2.28 (0.09) & 1.70 (0.11) \\ \bottomrule
    \end{tabular}
    \label{tab:essay-summary}
\end{table}

\subsection{Rule violation heatmaps}

\begin{figure}[ht]
    \centering
    \includegraphics[width=0.95\linewidth]{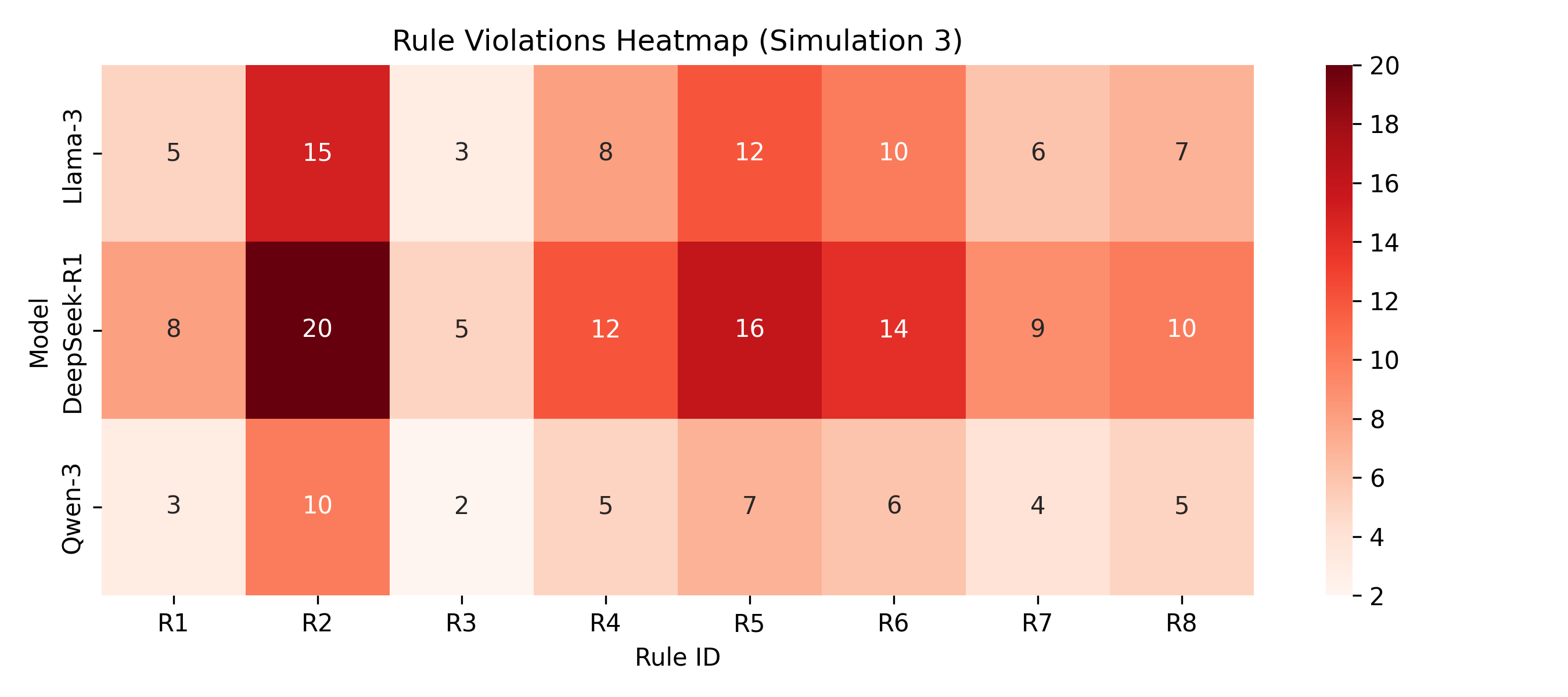}
    \caption{Heatmap of rule violations per model in the essay domain under simulation 3.  Darker cells indicate more frequent violations.  The patterns reflect model‑specific weaknesses and inform targeted red‑teaming.}
    \label{fig:heatmap}
\end{figure}

Figure\,\ref{fig:heatmap} visualises the distribution of rule violations across models in simulation 3.  DeepSeek‑R1 violated structural (R3, R4) and ethical (R5, R6) rules more often than other models, while Qwen‑3 exhibited the fewest violations.  Such heatmaps enable developers to pinpoint agent weaknesses and prioritise rule refinement.

\subsection{Enforcement action breakdown}

Table\,\ref{tab:enforcement} summarises enforcement actions across domains and simulations.  In the trading domain GaaS mostly issued blocks because financial rules were coercive, whereas in the essay domain warnings were more common.  Simulation 3 shows increased escalations due to adversarial injection.

\begin{table}[ht]
    \centering
    \caption{Breakdown of enforcement actions by domain and simulation.  Each cell reports the number of actions resulting in allow, warn or block.  Escalations refer to human review triggers.}
    \begin{tabular}{lccccc}
    \toprule
    Domain & Simulation & Allow & Warn & Block & Escalate \\ \midrule
    Essay & Sim 2 & 120 & 45 & 12 & 3 \\
    Essay & Sim 3 & 80 & 60 & 30 & 10 \\
    Trading & Sim 2 & 9 & 0 & 33 & 0 \\
    Trading & Sim 3 & 7 & 5 & 693 & 5 \\ \bottomrule
    \end{tabular}
    \label{tab:enforcement}
\end{table}

\subsection{Trust trajectories}

\begin{figure}[htbp]
    \centering
    \includegraphics[width=0.95\linewidth]{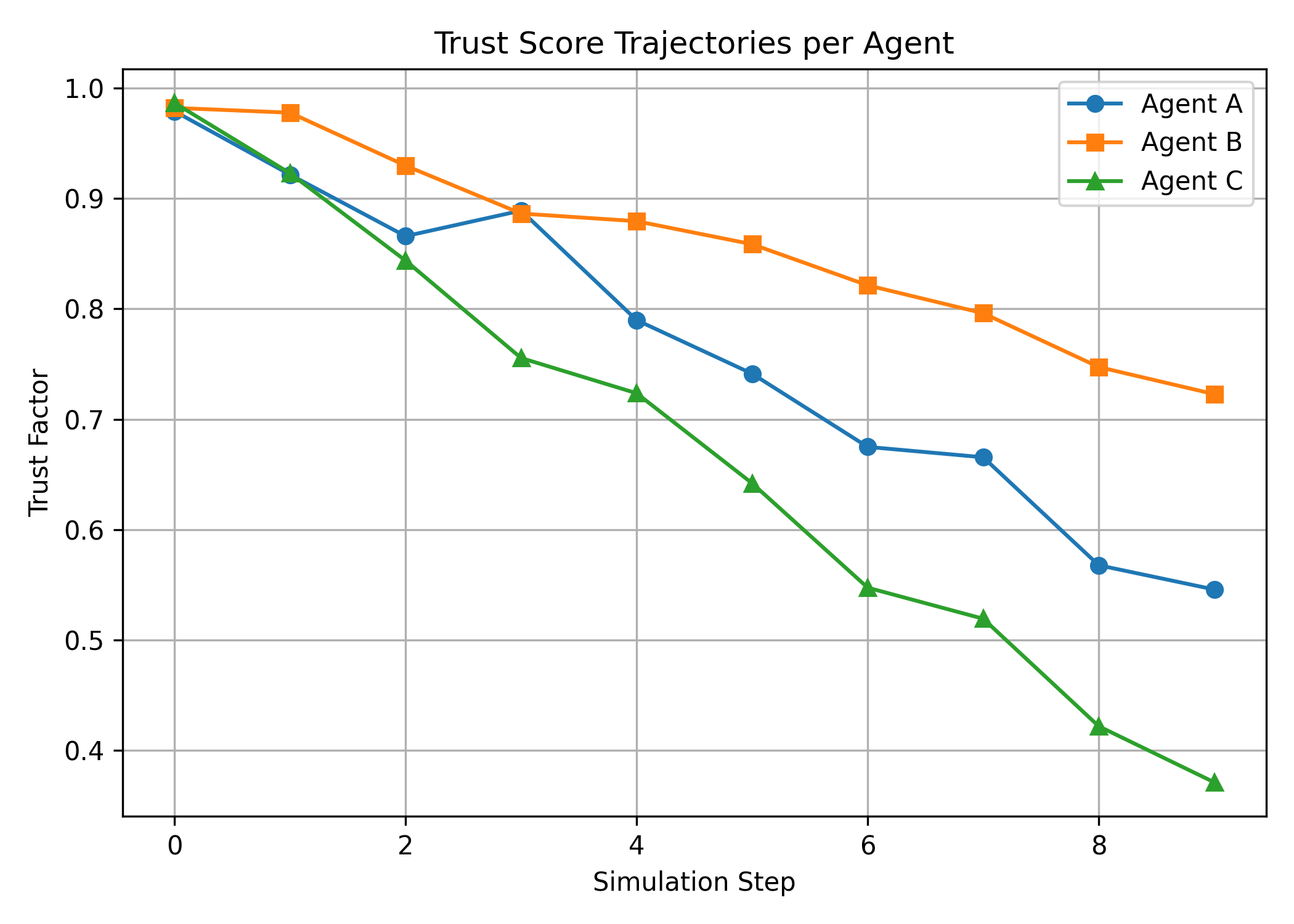}
    \caption{Trust score trajectories for three agents. Agent A incurs a mix of violations, Agent B remains mostly compliant, and Agent C commits repeated coercive violations. GaaS dynamically adjusts enforcement as trust declines.}
    \label{fig:trajectories}
\end{figure}

Figure\,\ref{fig:trajectories} plots trust factor trajectories for three representative agents across a ten‑step simulation.  Trust declines gradually for compliant agents (Agent B), whereas agents with repeated coercive violations (Agent C) experience rapid decay and eventual blocking.  Such trajectories can help system operators identify risk early and intervene.

\paragraph{Full logs.}  

The complete violation logs, trust trajectories and simulation outputs for all agents, rules and runs are provided as CSV files in the supplementary materials.  These datasets enable independent verification of the results and further analyses, such as rule‑specific false positive rates.

\end{document}